\documentclass[10pt,letterpaper]{article}

\usepackage{cogsci}

\cogscifinalcopy 

\usepackage{pslatex}
\usepackage{apacite}
\usepackage{float} 

\usepackage{natbib}

\usepackage{booktabs}
\usepackage{graphicx}
\usepackage[dvipsnames]{xcolor}

\newcommand{\citeposs}[2][]{\citeauthor{#2}'s (\citeyear[#1]{#2})}
\newcommand{\tuple}[1]{\ensuremath{\left\langle #1 \right\rangle}}

\definecolor{indianred}{RGB}{205,92,92}
\usepackage{nicefrac}

\usepackage{amsmath}
\DeclareMathOperator*{\argmaxA}{arg\,max} 

\title{Deep daxes: Mutual exclusivity arises through both learning biases and pragmatic strategies in neural networks}
 
\author{{\large \bf Kristina Gulordava (kgulordava@gmail.com)} \\
  \AND {\large \bf Thomas Brochhagen (thomasbrochhagen@gmail.com)} \\
  Universitat Pompeu Fabra, Barcelona, Spain\\
  \AND {\large \bf Gemma Boleda (gemma.boleda@upf.edu)} \\
  Universitat Pompeu Fabra, Barcelona, Spain / ICREA, Barcelona, Spain}

\begin{document}

\maketitle

\begin{abstract}

Children's tendency to associate novel words with novel referents has been taken to reflect a bias toward mutual exclusivity.
This tendency may be advantageous both as (1) an ad-hoc referent selection heuristic to single out referents lacking a label and as (2) an organizing principle of lexical acquisition. 
This paper investigates under which circumstances cross-situational neural models can come to exhibit analogous behavior to children, focusing on these two possibilities and their interaction. To this end, we evaluate neural networks' on both symbolic data and, as a first, on large-scale image data.
We find that constraints in both learning and selection can foster mutual exclusivity, as long as they put words in competition for lexical meaning. 
For computational models, these findings clarify the role of available options for better performance in tasks where mutual exclusivity is advantageous. For cognitive research, they highlight latent interactions between word learning, referent selection mechanisms, and the structure of stimuli of varying complexity: symbolic and visual.


\textbf{Keywords:} 
neural networks; mutual exclusivity; acquisition; pragmatics; learning biases; lexical meaning; referent selection
\end{abstract}

\section{Introduction}\label{sec:introduction}

A central puzzle in vocabulary acquisition is
how expressions come to be associated with meanings since, among manifold challenges, there are always multiple candidate referents for a novel word~\citep{quine:1960}. 
Prima facie, a learner that encounters, say, the word {\em rabbit} for the first time can entertain the hypothesis that it refers to any candidate meaning in the context of utterance; both with respect to a single referent (e.g., \textsc{rabbit}, \textsc{paw}, or \textsc{cuddly beast}) as well to others that might be salient in the scene (e.g., \textsc{table}, \textsc{boy}, or \textsc{feed}). However, children ultimately overcome this and other challenges faced during acquisition \citep{carey+bartlett:1978,bloom:2000}. 

A well-attested behavior in vocabulary acquisition is that learners (both children and adults) show a tendency toward \textbf{mutual exclusivity}. That is, they assume that a novel word refers to a novel object.
For instance, when prompted with an unknown word (e.g., ``show me the \emph{dax}'') and an array of familiar objects together with an unfamiliar object, children as young as $15$ months old tend to select the unfamiliar referent (\citealt{halberda:2003,markman+etal:2003}; see \citealt{lewis+etal:2019} for a recent review and meta-analysis). By contrast, recent work suggests that standard neural network (NN) models tend to associate novel input with frequent and familiar output. Crucially, they do so although mutual exclusivity would improve their task performance \citep{ghandi+lake:2019}.

In this study, we look at the conditions that make mutual exclusivity arise in a system that acquires a vocabulary in a challenging cross-situational setup. More specifically, we look at the interaction between \textbf{learning biases}, on the one hand, and \textbf{referent selection} strategies, on the other.
We focus on neural networks because they are powerful learning models that can scale to naturalistic input data such as audio and images. 
We analyze their performance on novel word comprehension, using tasks inspired by the ones children have been tested on \citep[e.g.,][]{horst+samuelson:2008}. Our main contributions are: (1) a systematic evaluation of how a NN's tendency to associate novel words with novel referents is impacted by its learning biases and its referent selection strategy; (2) a formalization of mutual exclusivity during reference selection in terms of Bayesian inference, highlighting ties to  probabilistic pragmatic models \citep[e.g.,][]{goodman+frank:2016,bohn+frank:2019}; and (3) evaluations on both symbolic and visual data that showcase how mutual exclusivity, as well as learning biases vs.\ referent selection strategies more broadly, interact with the structure of stimuli. To the best of our knowledge, this work is the first to study mutual exclusivity in large-scale image data with natural object co-occurrences. 

Our results show that mutual exclusivity  
can be fostered both during training, through a constraining loss function, or in on-line referent selection, through pragmatic-like reasoning. The core requirement is that there be a bias against synonymy in either realm (or both), making new words less likely to be associated with familiar referents. Considering not only symbolic but visual data additionally reveals that the success of this bias hinges on another requirement that may not always be fulfilled: objects need to be sufficiently discriminated.
This prerequisite may go unnoticed when evaluating models on symbolic datasets, or when conducting experiments with human subjects.

\section{Background}
 %
Children's tendency to select unfamiliar objects when prompted with unknown words has often been attributed to {\em mutual exclusivity} (ME; see, e.g., \citealt{markman+wachtel:1988,halberda:2003,markman+etal:2003,halberda:2006}). Pre-theoretically, ME can be characterized as a propensity to associate novel words with objects for which no linguistic label is known. It can be construed in different ways. First, it could be the result of a pragmatic referent selection strategy \citep[e.g.,][]{ clark:1988,halberda:2006,frank+etal:2009,bohn+frank:2019}. Intuitively, if an object has a known label, then a speaker should use this label to refer to it rather than a novel word. Consequently, a novel word can be reasoned to map, or at least circumstantially refer to, an unknown object. Second, it could be a learning bias, linked more closely to meaning acquisition and retention  (e.g., \citealt{markman+wachtel:1988, markman+etal:2003}; but see also \citealt{carey+bartlett:1978,horst+samuelson:2008, mcmurray+etal:2012} on meaning retention over time in ME tasks). For instance, it may be that already established word-meaning associations inhibit the linkage of new words to a meaning. 

Disentangling potential causes of an observed ME bias is not straightforward. Referent selection presupposes learning; and, inversely, latent learning biases cannot simply be read off from how children select referents. Ultimately, the interaction between these factors needs to be considered \citep{clark:1988,mcmurray+etal:2012}. In the following, we use {\em ME} as an umbrella term that is agnostic to the causes of the phenomenon, following \citet{lewis+etal:2019}, and hone in on these two potential causal factors and their interaction in computational models.

Previous models of cross-situational learning that evaluate novel word comprehension include probabilistic and connectionist approaches (e.g., \citealt{ichinco+etal:2008, frank+etal:2009,alishahi+etal:2008,fazly+etal:2010, mcmurray+etal:2012}; see \citealt{yang:2019} for a recent overview). However, these models are not easily scalable to large lexicons, nor can they directly ground word learning on more naturalistic data, e.g., images. By contrast, several scalable NN models that can learn word representations from aligned language-image data in cross-situational settings have been proposed \citep[e.g.,][]{synnaeve+etal:2014,lazaridou+etal:2016, chrupala+etal:2015}. None of these models, however, was evaluated on novel word reference with distractors: the classic setup in which children were put to the test.
More generally, little attention has been paid to the disentanglement of training biases and evaluation conditions in the success of word learning by NNs.
We explicitly focus on the consequences and desirability of mutual exclusivity as part of a network's training; as a part of its referent selection criterion; or as a combination thereof. 

Closer to our present efforts, \citet{ghandi+lake:2019} evaluate NNs on training-induced tendencies toward one-to-one mappings. Their findings suggest that common deep learning algorithms exhibit a learning behavior contrary to ME, tending to associate novel input to frequent and familiar output. This is a bad fit to machine learning tasks such as translation or classification. We largely share Gandhi \& Lake's motivations but focus on how ME can be brought about by interactions of training and referent selection criteria.
 

\section{Models}\label{sec:models}

Our goal is to study NNs behavior when prompted by a novel word as a function of word learning and referent selection strategies. We now introduce these two components.

\subsection{Component 1: Word learning}\label{sec:learning}

\begin{figure}
	\centering
\includegraphics[scale=0.25]{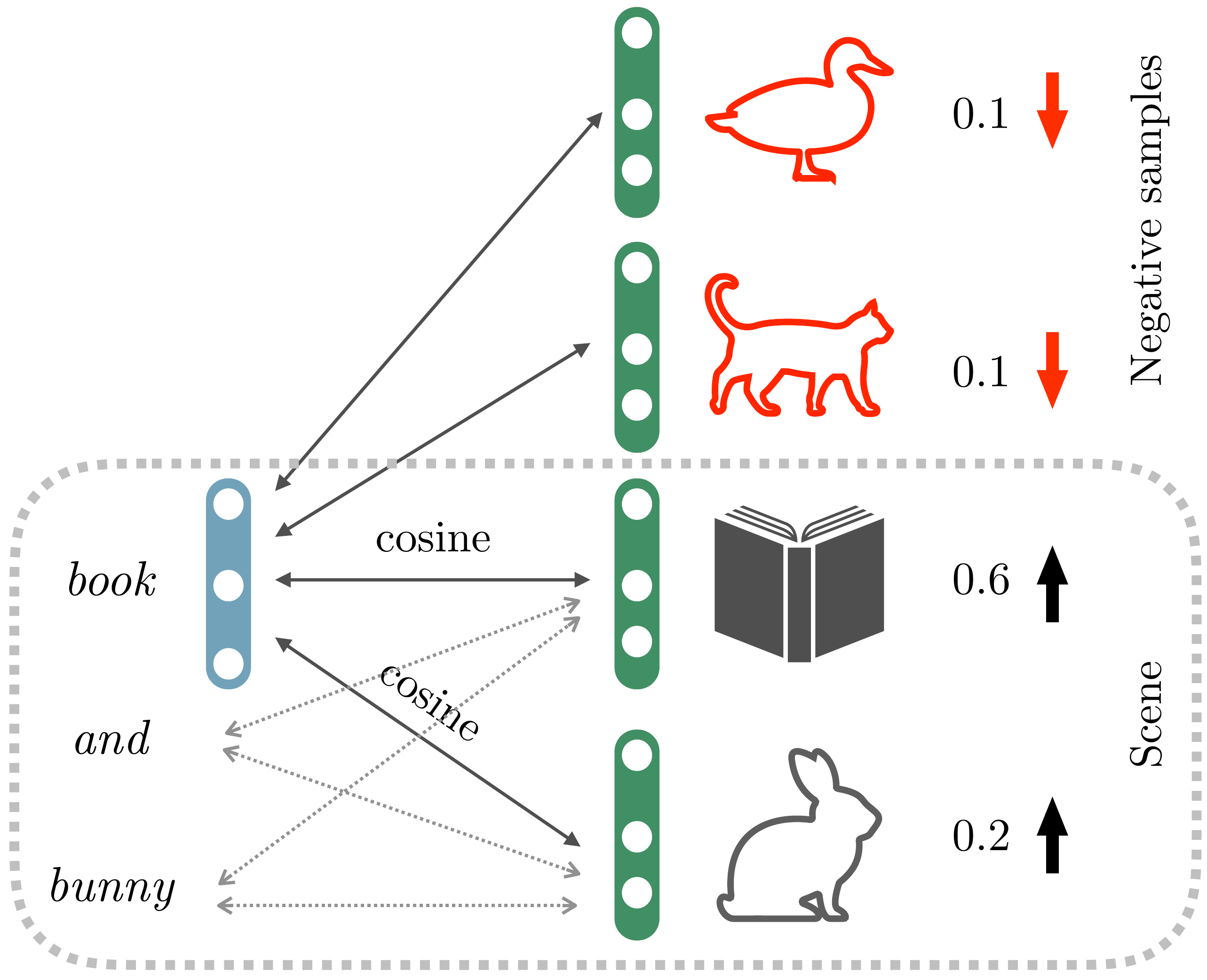}
\caption{Illustration of \textbf{max-margin loss over objects} for the utterance \emph{book and bunny} with objects \textsc{book} and \textsc{rabbit} in the scene. \textsc{cat} and \textsc{duck} exemplify negative examples for the positive input pair $\langle$\emph{book}, \textsc{book}$\rangle$.}
\label{fig:schemaloss}
\end{figure}

For our models, an input data point is a set of (potentially referring) words, $W$, and a set of objects in a scene, $S$ (see Figure~\ref{fig:schemaloss} for an example). We follow a common approach to train neural similarity models to simplify the optimization problem in cross-situational setups \citep[e.g.,][]{lazaridou+etal:2016}: we align all possible pairs of words and objects in the scene independently. 
The training input is then a
set of all word-object pairs, $\{\tuple{w , o} \mid w \in W , o \in S\}$. This is a simplification in the sense that the model is blind to the relation between objects in a shared scene.

Similarities between objects and words -- their learned association -- are computed from their encodings into a shared hidden space: 
\begin{equation}\label{eq:sim-model}
\mathbf{w} = E(w), \qquad 
\mathbf{o} = V(o), \qquad
sim = \cos(\mathbf{w}, \mathbf{o}), 
\end{equation}
where $E$ is an embedding of word $w$ and $V$ is a visual encoder of object $o$. In other words, the model in (\ref{eq:sim-model}) learns associations between words and objects, being fully parametrized by $E$ and $V$. In the experiments below, words and objects are encoded as $200$-dimensional vectors.


A similarity model like (\ref{eq:sim-model}) can be optimized in various ways. In particular, the values of $\cos(\mathbf{w}, \mathbf{o})$ will depend on a model's loss function objective. We implement three classes of objectives using max-margin loss.\footnote{A common alternative to max-margin loss is cross-entropy over softmax classification. We focus on max-margin loss because it is more generally applicable: it does not require a discrete vocabulary or set of objects. Experiments using softmax for the symbolic dataset yielded the same qualitative trends as those reported below. } They correspond to three major categories of learning constraints found in cross-situational word learning models. They either induce (a) competition among referents, imposing a soft ``a word maps only to one object''-constraint \cite[e.g.,][]{lazaridou+etal:2016,fazly+etal:2010}; or (b) competition among words, i.e., an ``an object maps only to one word''-constraint \cite[e.g.,][]{frank+etal:2009}; or they (c) induce competition over words and referents, equivalent to favoring one-to-one word-object mappings \citep[e.g.,][]{mcmurray+etal:2012,synnaeve+etal:2014}. Intuitively framed, (a) is an anti-polysemy bias, (b) is an anti-synonymy bias, and (c) is a combination of both.

\paragraph{(a) Anti-polysemy: Max-margin over objects}    
\begin{equation*}
  L_o = \sum_i \max(0, 1 - cos(\mathbf{w}, \mathbf{o})  + cos(\mathbf{w}, \mathbf{o_i})),
\end{equation*}
where object $o_i$ is a negative example, sampled randomly. While the similarity between the target word and the target object is increased, the similarity between the word and the negative example object is decreased. 

As illustrated in Figure~\ref{fig:schemaloss}, a positive example of {\em book} and its potential referents \textsc{book}, \textsc{rabbit} effects an increased association between them; whereas that of {\em book} and, e.g., negative referent examples \textsc{cat} and \textsc{duck} decreases. In other words, \textsc{book}, \textsc{rabbit}, \textsc{cat} and \textsc{duck} stand in competition for being associated with {\em book}. In the limit, negative sampling leads to competition among all objects.

Probabilistic models that learn $p(o \mid w)$ similarly enforce competition over objects: An increase in the probability of referent \textsc{book} given the word {\em book} decreases the probability of other objects ``competing'' for this name.

\paragraph{(b) Anti-synonymy: Max-margin over words}
  \begin{equation*}
      L_w = \sum_i \max(0, 1 - cos(\mathbf{w}, \mathbf{o})  + cos(\mathbf{w_i}, \mathbf{o})),
  \end{equation*}
    where word $w_i$ is a negative example, sampled randomly. In analogy to anti-polysemy, this leads to competition between words. While a positive example of the word {\em book} co-occurring with the referent \textsc{book} will increase their association, the association of \textsc{book} with, say, negative example words {\em kitty} and {\em duck} decreases. 
	


\paragraph{(c) One-to-one: Joint loss.}
Lastly, the combination of both losses implies competition over both words and objects, encouraging one-to-one-mappings: $L = L_w + L_o$


Note that, although alignments between novel referents and novel words are, by definition, never observed during training, we nevertheless allow for novel items --words or objects, depending on the loss objective-- to appear as negative examples. Otherwise, their relation to other items would solely be determined by their (typically random) initialization. Accuracy on referent selection, novel or not, presupposes a minimal degree of discriminability: among words and among objects. This is one way to encourage item discrimination even when not learning a particular association for them. We return to this issue below.

\subsection{Component 2: Referent selection}
To trace independent and joint effects of learning and referent selection criteria, we consider two selection mechanisms: Selection by only \emph{similarity} and by \emph{Bayesian inference}.

\paragraph{Referent selection as similarity match.} 
A straightforward strategy to pick a referent given a word $w$ is to always choose the one that most closely resembles the representation of the word. In other words, choosing the object with the highest similarity to the word out of all objects present in scene $S$:
\begin{equation}
  o^* = \argmaxA_{o \in S} cos(\mathbf{w}, \mathbf{o}). \label{eq:best-match}
\end{equation}
In probabilistic terms this is equivalent to choosing the object in the scene that maximizes $p(o \mid w,S)$. 


\paragraph{Referent selection as Bayesian inference.}
Our second criterion is in the spirit of pragmatic reasoning \citep{goodman+frank:2016,bohn+frank:2019}. The view of ME as such a referent selection criterion was prominently put forward by \citet{halberda:2006} in terms of a {\em disjunctive syllogism}, based on eye-tracking data that suggests that adults and preschoolers ``reject'' known referents before resolving novel word reference. Intuitively, one can can reason that, if the speaker intended to refer to an object with a known name, she would have used that name. Since she did not, but instead used a novel label, she must mean the unfamiliar object. 

More generally, the idea of modeling interpretation as an inversion of production has made much explanatory headway at the interface of experimental pragmatics and cognitive modeling (see \citealt{goodman+frank:2016} for review). However, these models have mainly been applied to small and discrete symbolic domains \citep[though see][]{andreas+klein:2016, monroe+etal:2017, zarriess+schlangen:2019}.\footnote{In particular, \citet{zarriess+schlangen:2019} also use Bayesian pragmatics in the context of novel word reference in complex scenes with natural images. By contrast, however, they focus on the generation of referring expressions for unseen objects, modeling speakers' probabilities using listeners' beliefs.}

\begin{figure}
	\centering
\includegraphics[scale=0.29]{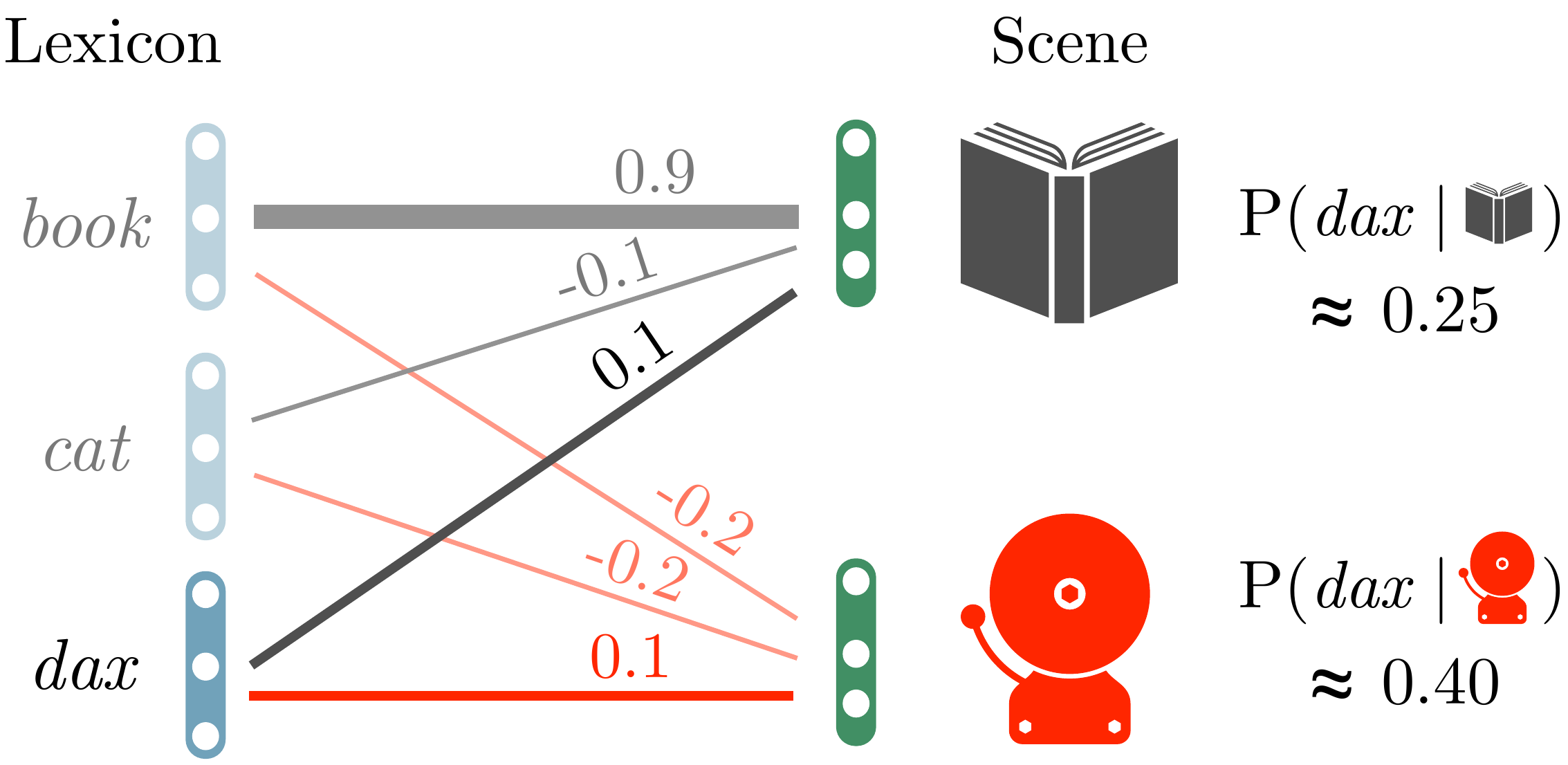}
\caption{Illustration of the relationship between novel word \emph{dax} and two objects in a scene, \textsc{book} and \textsc{dax}. Values linking words and objects are cosine-similarities. A model using the matching strategy ties between the two objects. A model using the Bayesian strategy picks \textsc{dax}.}
\label{fig:bayesianselection}
\end{figure}

The choice of the most likely referent from all objects in a scene given word $w$ can be written as $\argmaxA_{o \in S} p(o \mid w, S)$. In probabilistic pragmatic models, $p(o \mid w, S)$ is construed as the likelihood of a listener interpreting $w$ as $o$ in $S$. Due to the sparsity of actual observations for all potential scenes $S$, such listener probabilities are hard to model computationally. However, we can get at them through their inverse, the speaker probability, using Bayes' rule:
\begin{align}
  p(o \mid w, S) \propto p(w \mid o, S) p(o \mid S) \label{eq:bayes}
\end{align}
We make two assumptions to approximate the left-hand expression. First, we assume that the label used for an object depends only on itself and not on others present in a scene. This is a simplification. Linguistic choice can certainly vary as a function of other objects present in a scene. For instance, in a scene with two dogs, of which one is a Rottweiler, a speaker may prefer to say {\em Rottweiler} instead of {\em dog}  \citep[e.g.,][]{ferreira+etal:2005}. Second, we assume the prior over objects in a scene to be uniform. In naturalistic scenarios, speaker goals and contextual saliency can certainly skew this distribution. However, this assumption minimally holds true for the experimental conditions in which children are often tested in (see \citealt{brochhagen:thesis}:\S3.2 for discussion). With these provisos, the right-hand side of (\ref{eq:bayes}) simplifies to $p(w \mid o)$. Referent selection using Bayesian inference can then be rewritten as:
\begin{equation}
  o^* = \argmaxA_{o \in S} p(w \mid o). \label{eq:best-bayes}
\end{equation}
The speaker probability $p(w \mid o)$ is obtained from the similarity values learned by our neural models, normalizing $cos(\mathbf{w}, \mathbf{o})$ over all words in the vocabulary given object $o$. This is illustrated in Figure~\ref{fig:bayesianselection}.

Previous models have also exploited $p(w \mid o)$ to model referent selection tasks \citep[in particular][]{alishahi+etal:2008, fazly+etal:2010}. However, the motivation to do so in the context of reference to novel words was informal \citep[pp.~1045--6]{fazly+etal:2010}. By contrast, we just provided an explicit derivation of mutual exclusivity, construed as a referent selection criterion, using Bayesian inference. This clarifies how $p(w \mid o)$ is related to ME, building on theory-driven probabilistic pragmatic models.

To recapitulate, as shown in Figure~\ref{fig:bayesianselection}, referent selection as similarity match, in (\ref{eq:best-match}), picks the referent that most closely resembles the representation of a given word in a scene. Referent selection as Bayesian inference, in (\ref{eq:best-bayes}), additionally factors in alternative words that could have been uttered to refer to each object.

\section{Experiments}\label{sec:experiments}
We evaluate model performance of all the learning-selection combinations introduced above: Models are trained with max-margin loss over objects, words, or both; and select referents either by similarity or Bayesian inference.\footnote{Hyperparameters were determined using random search over a set of learning rates; initialization ranges for word and object embeddings; and hidden dimension sizes. We evaluate models at their lowest loss after a maximum of 20 epochs. One could worry that evaluating models at their least loss could lead to overfitting. However, note that optimizing the loss function does not imply optimizing accuracy on referent selection. Moreover, training includes no positive examples of alignments between novel items.}  
In analogy to experiments with children, test scenes for novel referent selection include one novel object and several familiar ones. Task success is defined as picking novel referents when prompted by novel words. 

We evaluate on two datasets. The first is a symbolic dataset of annotated child directed speech \citep{frank+etal:2009}. The second is a visual dataset, comprising images and associated captions  \citep{flickr30k-entities}. 



\subsection{Symbolic dataset}\label{sec:symbolic}

\paragraph{Data.} \citeposs{frank+etal:2009} data comes from two transcribed recordings from CHILDES \citep{childes}. It comprises $620$ utterances with around $3800$ tokens. Utterances are annotated with objects present at speech time, e.g., $W$ = \{\emph{get}, \emph{the}, \emph{piggie}\} and $S$ = \{\textsc{pig}, \textsc{cow}\}, respectively, for words and objects in a scene. A gold lexicon provides the correct alignments between $22$ objects and $36$ word types. 
There is a mean of $4.1$ words and $2.4$ objects per scene.

\begin{figure}[t]
 \centering
  \includegraphics[scale=0.35]{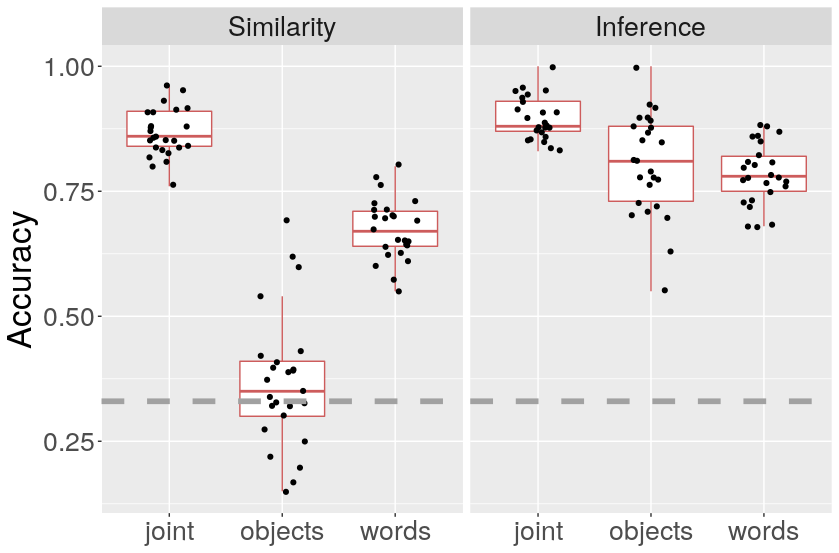}
 \caption{Mean accuracy on novel referents in $25$ experiments per condition, each independently initialized, with random baseline ($.33$) as dashed line. Means, from left to right: $[.87,.37, .67]$ and $[.9, .8, .78]$.} \label{fig:nwsymb}
 \end{figure}

\paragraph{Evaluation setup.} 
For familiar word comprehension, we report Best F-scores between the gold lexicon and the one learned by the models \citep[cf.,][]{frank+etal:2009,lazaridou+etal:2016}. Since F-scores are computed at type level we weight the loss computed for each target token by its inverse frequency in the corpus. 
To test performance on novel items, we added five novel words to the vocabulary (\emph{dax1}, ... , \emph{dax5}). Accuracy scores then come from evaluations of $20$ test scenes per novel word. Test scenes include one novel object and two uniformly sampled ones from training. For example, novel word {\em dax1} may be evaluated in scene \{\textsc{cat}, \textsc{cow}, \textsc{dax1}\}. 
\setlength{\tabcolsep}{2.9pt}
\begin{table}
  \center
\begin{tabular}{lr}
\toprule
          Loss &   Best F\\
\midrule
  joint&  .68 \footnotesize{(.03)} \\  
   over objects & .71 \footnotesize{(.01)}  \\
   over words & .65 \footnotesize{(.02)} \\
  \midrule
  Frank et al. 2009 & .55 \\
 Lazaridou 2016 \emph{--visual} (shuffled) & .65 \\
  Lazaridou 2016 \emph{+visual} & .70 \\
\bottomrule
\end{tabular}
\caption{Familiar word comprehension for CHILDES data: \textbf{Best F-scores} (mean and standard deviations) for learnt vs.\ gold lexicon  in $25$ experiments, each independently initialized. 
}
\label{t:bestf}
\end{table}
\paragraph{Results.} As shown in Table~\ref{t:bestf}, models achieve very good Best F-scores. They are close to the scores of \citet{lazaridou+etal:2016} although we do not consider relations between words nor additional visual input. However, as shown by the accuracies on referent selection with novel items in Figure~\ref{fig:nwsymb}, success at acquiring the lexicon is not indicative of performance on novel referent selection. If competition over words is encouraged during learning then picking by similarity is a viable, though suboptimal, selection strategy. Without this learning bias, picking by similarity results in random performance. Bayesian inference always presents an improvement over picking by similarity only, but this improvement's magnitude hinges on the loss' objective. A comparison of training with max-margin loss over words against one over objects shows that, if competition over words is enforced through a referent selection mechanism, then learning with a complementary bias against polysemy can be as or even more useful than imposing the anti-synonymy constraint in both training and selection.  
%

\subsection{Visual dataset: Flickr30k Entities}

\paragraph{Data.} Flickr30k Entities \citep{flickr30k-entities} contains images with crowd-sourced descriptions, and bounding boxes linking objects in images to their referring expressions. We pre-process this data to extract word-object annotations. For each referring expression in a caption, e.g., \emph{a smiling person}, we take the last word (\emph{person}) as the linked object's label.\footnote{Referring expressions with prepositional phrases, e.g.,  \emph{[a smiling person] with [Mohawk hairstyle]}, are annotated as two separate referring expressions aligned with different image regions.} The visual features of each object (bounding box) are then pre-computed using a convolutional NN VGG16 model trained on ImageNet \citep{simonyan+zisserman:2015}. We process them scaled to $224 \times 224$ pixels and take the output of the last layer of the model.

Each image is treated as a scene. This yields a natural co-occurrence distribution of objects. There are some important differences to the previous dataset. First, one word can refer to many instances of the same concept represented as different objects across images. The symbolic dataset did not have this distinction between instances and object categories, as it mapped all word uses to the same symbol (e.g., ``piggie'' always referred to \textsc{PIG}, regardless of whether different pigs where referred to in different scenes). 
Second, object embeddings are initially determined by the pre-processed VGG16 vectors rather than, as previously done, randomly initialized; this can be seen as analogous to assuming that children know how to visually represent objects in experimental conditions. Third, images can have up to five different descriptions. We treat the resulting word-object alignments as independent data points.

We excluded objects that span more than one bounding box, typically referred to by plurals, as well as cases in which one bounding box contained another one. This results in $29782$ images, a vocabulary of $6165$ referring words, and a total of $130327$ data points, with a mean of about $2.22$ objects per scene.

\begin{figure}[t]
 \centering
  \includegraphics[scale=0.52]{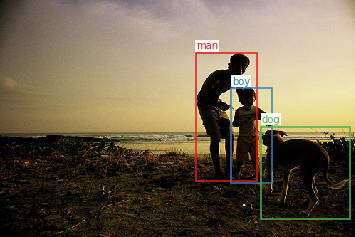}
 \caption{Test item (labels are for illustration only).} \label{fig:flickr}
 \end{figure}

\paragraph{Evaluation setup.} 
As there is no gold lexicon to benchmark against, we focus on model accuracy on referent selection when tested with both familiar and novel words. Images containing only one referent were excluded to avoid trivial solutions. We let dogs be a surrogate category for novel items: Where children would see unfamiliar objects in an otherwise familiar array and be prompted with a novel word, our models are trained without encountering positive examples of dogs nor of words used to refer to them (e.g., {\em dog, dogs, puppy, retriever, shepherd, corgi, pug, collie} or {\em spaniel}). We then evaluate the models on scenes containing dogs, as illustrated in Figure~\ref{fig:flickr}.
The particulars of this setup and the choice of a category as a stand in for a novel one certainly affect the results that follow. Our choices are motivated by wanting to retain the integrity of images as natural scenes; and by dogs appearing frequently enough in the data to ensure that a variety of different kinds of scenes with different numbers of objects are evaluated.

\paragraph{Results.} Table~\ref{tab:svisual} shows results for familiar word comprehension. Similar to the symbolic case, performance is well above random (0.42)
but not optimal. The low deviation across experiments suggests that all models have comparable endpoints, with models learning with anti-synonymy slightly outperforming those with anti-polysemy. Different ways of selecting referents did not impact accuracy for familiar items. Acquired word-referent associations are refined enough after sufficient training, leaving no room for Bayesian inference to further improve on them.
\begin{table}
  \center
\begin{tabular}{lcc}
\toprule
Loss &    Similarity &  Inference\\
\midrule
   joint & .64  & .64 \\
     over objects & .62 & .62 \\
  over words &  .65 & .65 \\
\bottomrule
\end{tabular}
\caption{Familiar word comprehension for Flickr30k Entities: Mean \textbf{accuracy} in $30$ experiments, each independently initialized (random baseline: $.42$; $\text{SD} < .009$ in all conditions).}
\label{tab:svisual}
\end{table}

\begin{figure}[t]
 \centering
  \includegraphics[scale=0.33]{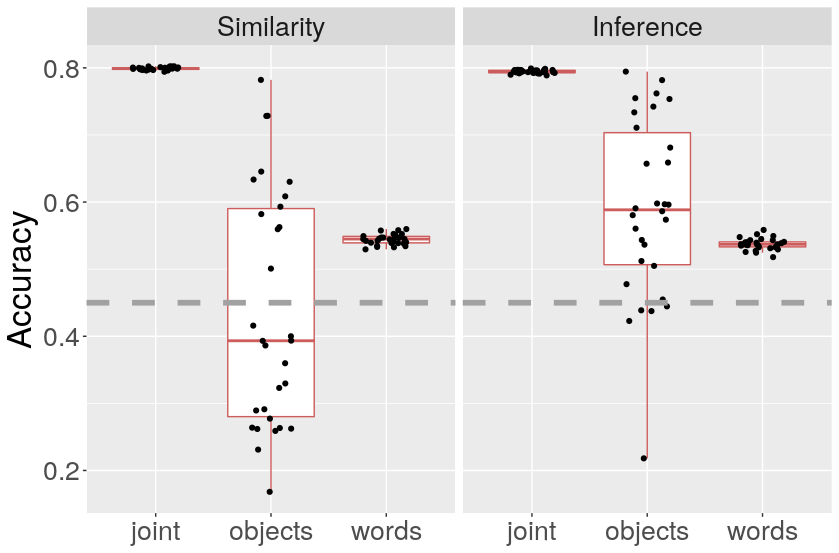}
 \caption{Mean accuracy on novel referents in $30$ experiments per condition, each independently initialized, with random baseline ($.45$) as dashed line. Means, from left to right: $[.8, .44, .54]$ and $[.79, .59, .54]$.} \label{fig:nwvisual}
 \end{figure}

As shown in Figure~\ref{fig:nwvisual}, things are different for novel items. Akin to the symbolic case, Bayesian inference offers an advantage to models trained with only loss over objects. When choosing by similarity only, performance is about random with only an anti-polysemy learning bias. There are two main differences from the symbolic case, however. First, Bayesian inference confers no advantage to models that learned with anti-synonymy. By contrast, it did provide a small edge on symbolic data. Second, in the visual case, anti-polysemy can be helpful. This is suggested by both the better performance of models learning with max-margin loss over objects that use Bayesian inference over models learning with only max-margin loss over words; as well as by the success of models trained with joint objectives compared to those trained only with anti-synonymy. More succinctly put, while max-margin loss over objects conferred no clear advantage in symbolic experiments, it did so in the visual ones.

To understand the positive effect of anti-polysemy for this set of experiments, let us first address another result particular to them: the contrast of deviances across max-margin objectives. This difference can be traced back to the consequences of the refinements they lead to. Max-margin loss over objects aligns a positive example of a word-object pair while separating this word from negative object examples (Figure~\ref{fig:schemaloss}). That is, referents seen as positive examples and referents seen as negative ones are pulled apart. This loss objective thus leads to improved object discriminability. Since word discriminability is not directly improved upon, however, the performance of models with only an anti-polysemy bias is sensitive to the random initialization of (novel) word embeddings. This leads to large variations across experiments. By contrast, since visual embeddings were not initialized randomly but pre-trained, no such deviance is observed for losses that improve only word discrimination. The answer to the question of what can make anti-polysemy advantageous is then that it improves object discriminability; the amount of objects and their (visual) resemblance being a major difference between our datasets. Nevertheless, learning to discriminate words remains pivotal to our task. As a consequence, the joint objective, profiting from both increased object discriminability and increased word discriminability, outperforms either individual loss objective.      

%
 
\section{Conclusion}\label{sec:conclusion}
We have shown that mutual exclusivity, and an ensuing success on novel word comprehension, can be achieved with scalable models with continuous representations and conventional learning algorithms (contra, e.g., \citealt{ghandi+lake:2019}). For this to happen, competition over words needs to be induced: either during learning, through a constraining loss objective, or during referent selection, through pragmatic reasoning. This requirement mirrors broader patterns found in natural language: While the existence of true synonyms is contested, there is little doubt about the abundance of polysemy \citep{brochhagen:thesis,clics3}. Although, in principle, anti-polysemy is not required for success on ME our results on visual data paint a nuanced picture. While anti-synonymy alone can lead to moderate success on this difficult task, learning biases that encourage task-specific discrimination of objects (here: visually) can further improve on it. One way to encourage such discrimination is through negative examples, as done here. Another is to manipulate item initialization as a function of the task and data, akin to having special ``slots'' for novel items. We hope that future work will address the specifics of such a manipulation and its comparison with the kind of acquired discrimination we have investigated here. More broadly, our results highlight the importance of evaluating word learning models on more complex and varied datasets, since trends observed on small symbolic data do not necessarily scale up to visual features and large lexica.\footnote{Concurrent work by \citet{vong2020learning}, analyzing word learning from raw images of digits using NNs, exemplifies growing efforts in this direction.} 



\section{Acknowledgments}
We gratefully acknowledge the AMORE team for feedback, advice and support. This project has received funding from the European Research Council (ERC) under the European Union's Horizon 2020 research and innovation programme (grant agreement No 715154), and from the Spanish Ram\'on y Cajal programme (grant RYC-2015-18907).
We thankfully acknowledge the computer resources at CTE-POWER and the technical support provided by Barcelona Supercomputing Center (RES-IM-2019-3-0006). We are grateful to the NVIDIA Corporation for the donation of GPUs used for this research. We are also very grateful to the Pytorch developers. This paper reflects the authors' view only, and the EU is not responsible for any use that may be made of the information it contains.\\ \includegraphics[width=2cm]{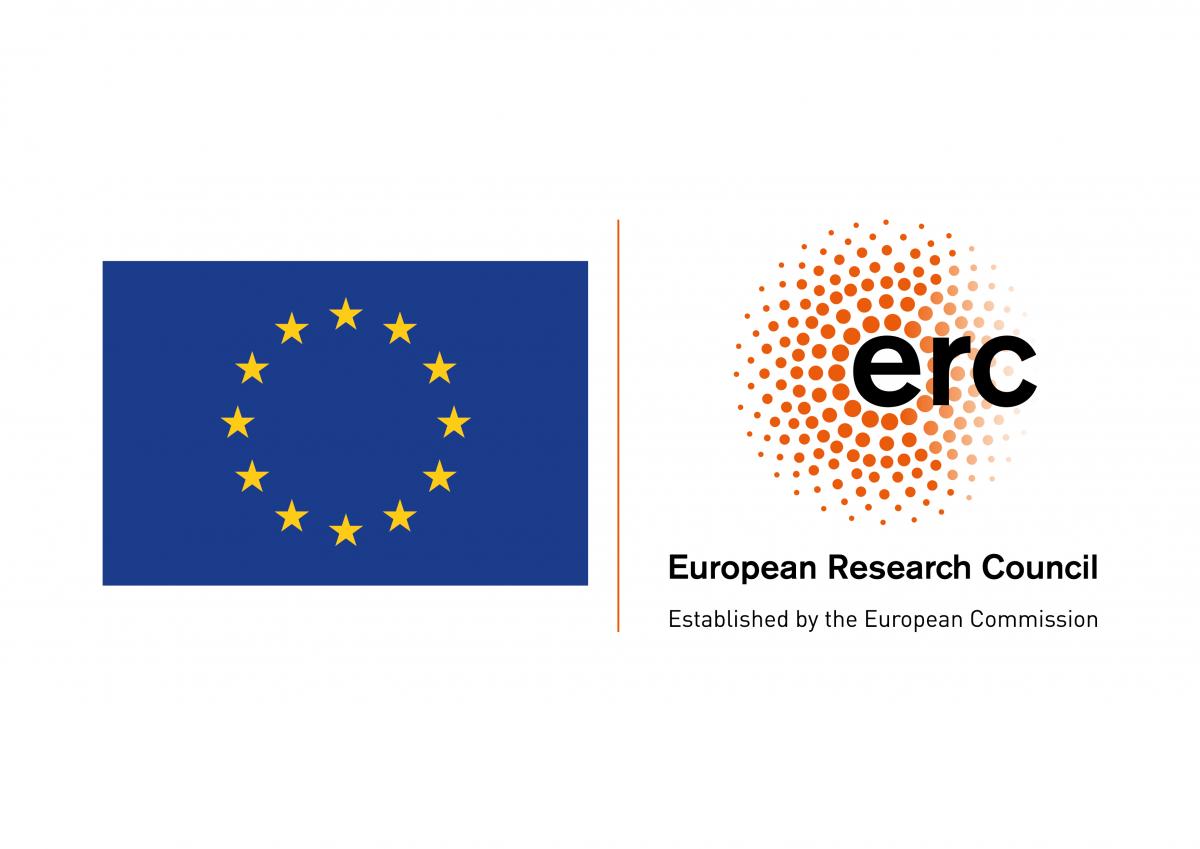} 

\bibliographystyle{apacite}

\setlength{\bibleftmargin}{.125in}
\setlength{\bibindent}{-\bibleftmargin}

\bibliography{me}

\end{document}